Bored to Death: Artificial Intelligence Research Reveals the Role of Boredom in Suicide Behavior

Accepted manuscript before proofing.

*Frontiers in Psychiatry*[a]

Shir Lissak[1]

Yaakov Ophir, PhD[1,2]

Refael Tikochinski[1]

Anat Brunstein Klomek, PhD[3]

Itay Sisso[4]

Eyal Fruchter, MD[1]

Roi Reichart, PhD[1]

[1]Technion – Israel Institute of Technology, [2]University of Cambridge [3]Reichman University, [4]The Hebrew University of Jerusalem

Correspondence:  yaakovophir@gmail.com

## Abstract

**Background:** Recent advancements in Artificial Intelligence (AI) contributed significantly to suicide assessment, however, our theoretical understanding of this complex behavior is still limited.

**Objective:** This study aimed to harness AI methodologies to uncover hidden risk factors that trigger or aggravate suicide behaviors. **Methods**: The primary dataset included 228,052 Facebook postings by 1,006 users who completed the gold-standard Columbia Suicide Severity Rating Scale. This dataset was analyzed using a bottom-up research pipeline without a-priory hypotheses and its findings were validated using a top-down analysis of a new dataset. This secondary dataset included responses by 1,062 participants to the same suicide scale as well as to well-validated scales measuring depression and boredom. **Results:** An almost fully automated, AI-guided research pipeline resulted in four Facebook topics that predicted the risk of suicide, of which the strongest predictor was boredom. A comprehensive literature review using *APA PsycInfo* revealed that boredom is rarely perceived as a unique risk factor of suicide. A complementing top-down path analysis of the secondary dataset uncovered an indirect relationship between boredom and suicide, which was mediated by depression. An equivalent mediated relationship was observed in the primary Facebook dataset as well. However, here, a direct relationship between boredom and suicide risk was also observed. **Conclusions:** Integrating AI methods allowed the discovery of an under-researched risk factor of suicide. The study signals boredom as a maladaptive 'ingredient' that might trigger suicide behaviors, regardless of depression. Further studies are recommended to direct clinicians' attention to this burdening, and sometimes existential experience.

**Keywords:** Boredom; Suicide Prevention; Social Media; Risk Factors; Large Language Model (LLM); Artificial Intelligence (AI)

## Introduction

Suicide, one of the major public-health concerns today following the COVID-19 crisis,[1-3] is a highly complex human tragedy.[4] Unfortunately, despite decades of research, our understanding of this somewhat enigmatic phenomenon is unsatisfying, as implied in the well-cited meta-analysis by Franklin et al.[5] According to this wide-scope meta-analysis, "*the [suicide] field has primarily focused on the same risk factors for the past 50 years, with risk factors becoming increasingly homogenous over past five decades*". Nevertheless, a significant change may occur today, with the sweeping revolution in Artificial Intelligence (AI) and the ubiquitous spread of social media.[6]

The recent introduction of strong Large Language Models (LLMs), such as GPT-3,[7] allowed researchers to analyze large amounts of authentic and valuable personal texts, which became highly accessible with the growing popularity of the various social networking sites (e.g., Facebook, Twitter). LLMs are capable of capturing subtle and meaningful patterns in raw data (e.g., posts or tweets) and previous research had already demonstrated their promising potential for suicide predictions.[8] To date, dozens of AI studies reported of high quality predictions of suicide risk from social media, using purely bottom-up methods, which did not involve the more traditional, top-down examination of pre-defined hypotheses.[9] The few that do include top-down examination require pre-defined hypotheses and in that manner do not allow new scientific discoveries by themselves.[10] Notably, bottom-up AI studies often achieved superior predictions than the pure top-down, theory-driven studies,[11-13] however, to our knowledge, these AI-based predictions were never translated into actual theoretical advancements in suicidology.

In the absence of pre-defined hypotheses about specific risk factors and the opaque nature of the LLMs, which are often referred to as 'black box' models,[14] it has been difficult to pinpoint the exact patterns or themes (i.e., risk factors) that drove these complex models to make their high quality predictions. In a way, the improved predictions generated by the LLMs came at a cost, as we now struggle to understand the internal mechanism of these models. As opposed to the top-down studies that typically examined a limited number of well-defined risk factors, the bottom-up AI studies analyzed numerous and implicit, data-driven features, thus restricting our ability to isolate the concrete psychosocial risks that might have been involved in the creation or maintenance of the suicide behaviors.

The overall goal of this research was to address this gap and harness the power of the LLMs for scientific discovery of risk factors. To exhaust the full potential of the LLMs, the first steps of the research pipeline were designed in a purely bottom-up manner, so that the results will not reflect upon predefined risk factors, but on authentic, data-driven, and perhaps less researched factors. Notably, the results from these almost fully automated steps indicated that the topic (i.e., theme) that contributed the most to the prediction of suicide addressed *boredom* experiences.

Boredom, or the "unfulfilled desire to be engaged in satisfying activity" as defined by Fahlman and collegues,[15] is typically accompanied by mild negative emotions, low arousal and attentiveness, and decreased sense of meaning in life.[16] This negative experience of boredom is even considered as a significant risk factor, or an inherent component, of depression.[17,18] However, to our knowledge, the role of boredom in the emergence and maintenance of suicide ideation and behaviors has not been characterized in the literature. In fact, following the 'bottom-up' discovery of boredom, we conducted a literature review using *APA PsycInfo*, the abstracting and indexing database of the American Psychological Association, and found that boredom is rarely perceived as a unique risk factor of suicide (for more information, see the Discussion section).[19-22]

We therefore completed the research pipeline with a final step that consisted of the collection of a new dataset. Using this new dataset, we could now examine a pre-defined hypothesis (that emerged from the previous bottom-up steps) that boredom experiences will predict suicide risk, either directly or indirectly, through the mediating variable of depression. In this way, we could further characterize the role of boredom in suicide behavior and illustrate how the new LLMs can be leveraged, not only for practical prediction of suicide risk, but also for theoretical advancements in suicidology (Discussion).

## Method

### Data

The collection of the data was conducted with the ethical approval of the Institutional Review Boards of the Hebrew University of Jerusalem and the Technion – Israel Institute of Technology. The complete description of the primary dataset of the current study is available in our previous publication that focused on prediction (rather than on understanding) of suicide.[9] Briefly, upon reading and signing a consent form, participants recruited from Amazon's Mechanical Turk (MTurk) completed common psychiatric and psychological questionnaires and gave a one-off authorization to download their Facebook activity up to 12 months prior to the research date. This activity was extracted to an encrypted data storage through a designated application that was developed for this purpose.

Altogether, the primary dataset consisted of 228,052 Facebook postings that were uploaded by 1,006 active Facebook users (23.26% male). The participants were English speaking residents of the US (mean age = 44.7, $SD$ = 13.9). The quality of the participants' responses was ensured via subtle measures and attention checks we developed (for further information about this primary dataset, see the Supplementary Material).[23]

Importantly, the collected postings were matched to the users' responses to the well-researched Columbia Suicide Severity Rating Scale (CSSRS),[24] which was administered in the original study. The CSSRS is considered a 'diagnostic tool of choice', both in clinical settings and in empirical research.[25,26] Upon consultation with the principal developer of the CSSRS (Posner, personal written communication), we chose to administer the electronic, screening version of the scale, in light of the fact that the research examined participants from a crowdsourcing platform. This version has been demonstrated to have psychometric validity and prediction accuracies that are comparable to the original clinician version of the scale.[27,28]

The structure of the CSSRS contains two parts. In the first part (items 1-2), participants are asked to respond YES or NO to questions about passive suicide ideation. In the second part (items 3-6, which are only shown to the participants if the first part indicated they had thoughts of killing themselves), the participants are asked about active suicide ideation (i.e., suicidal thoughts with method, intent, or a specific plan) suicide behavior (i.e., real-life activities aimed at ending one's life, such as collecting pills or obtaining a gun). In this study, the distribution of the participants' scores was as follows: 64.01% received zero, 10.47% received 1, 12.36% received 2, 6.08% received 3, 2.99% received 4, 3.29% received 5, and 0.8% received 6.

**Procedures (a 5-step research pipeline)**

Based on this high-quality dataset, we designed a 5-step research pipeline (Figure 1). A complete description of all the methodological details of this pipeline is provided in the Supplementary Material. The description below, in the current section, provides a brief overview of the five consecutive procedures that comprised the research pipeline, including the key computational methods that were implemented in each step and the secondary dataset that was collected in the fifth step.

Figure 1

*An overall representation of the research pipeline*

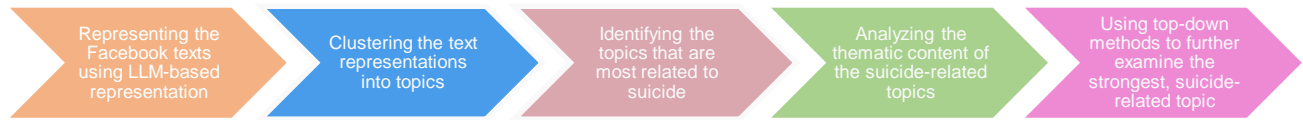

First, we used a state-of-the-art (yet still "black box") LLM named SBERT[29] to assign vector representations to the Facebook postings. Second, we applied a clustering algorithm named HDBSCAN,[30] with the goal of organizing the different posts (represented numerically) into groups that ideally capture meaningful 'topics' (i.e., themes). Third, we conducted a stepwise regression[31] and identified the topics that contributed the most to the prediction of suicide (i.e., the CSSRS scores). Fourth, we analyzed the thematic content of the resulted topics using three methods: manual inspection of representative postings, 'consultation' with ChatGPT,[32] and content analysis using the well-established method of TF-IDF (Term Frequency – Inverse Document Frequency).[33]

Fifth, to validate and further examine the bottom-up results from the previous steps, which suggested that suicide risk is linked to boredom experiences (Results), a secondary new sample of 1,062 participants was collected (52% male, mean age = 44.7, *SD* = 13.9). In this *secondary, questionnaire-based dataset*, participants completed three well-validated psychological measurements: (1) the aforementioned suicide scale,[24] (2) the 9-item Patient Health Questionnaire (PHQ-9) that is commonly used to measure depression,[34-36] (considering its potential mediating role that was described in the Introduction), and (3) a well-researched measurement of boredom named the Multidimensional State Boredom Scale (MSBS).[15]

The descriptive information about the CSSRS is presented above. The PHQ-9 is a nine-item scale that targets the nine symptoms of depression that appear in the fifth edition of the Diagnostic and Statistical Manual of Mental Disorders (DSM-5).[37] Following the DSM diagnostic criteria, a cut-off point for the presence of depression can be calculated when participants report of five or more symptoms that disturbed them in the past two weeks, for at least "more than half the days". In addition, at least one of these symptoms should refer to either low interest or depressed mood, which are the core symptoms of the disorder.[34] In the current secondary sample, a total of 130 users (12.24%) met this DSM-based cut-off point for depression.

The MSBS consists of 28 items designed to evaluate the state of boredom. An additional scale measuring the more general tendency of the person to feel bored (i.e., the Boredom Proneness Scale)[17] was also implemented in this secondary sample (see the Supplementary Material). However, the following analyses relate to the *state* boredom scale only to match the two additional scales of the study that evaluated the current/recent (state) depressive and suicidal experiences of the participants (i.e., the PHQ-9 and the CSSRS). The state boredom scale also reflected best the boredom-related expressions that were identified in the primary Facebook data.

The 28 items of the MSBS are classified into five theoretical dimensions: Disengagement, High arousal, Low arousal, Inattention, and Time perception. Of particular interest to our investigation is the central dimension of Disengagement. This dimension, which consists of the largest number of items ($N = 10$) and has the highest factor loading (0.97) includes items, such as: "Everything seems repetitive and routine to me", "I feel bored", and "I want to do something fun, but nothing appeals to me".[15] The mean scores of the participants on this dimension of Disengagement was 2.26 ($SD = 1.56$, range = 0 to 6).

Notably, this Disengagement dimension was found to be the closest dimension (out of the five potential dimensions of boredom) to the boredom expressions that were common in the primary Facebook data. Various similarity analyses, which are described in the Supplementary Material ensured this similarity formal concept of boredom, as manifested in this Disengagement dimension, and the informal, Facebook-based topic of boredom. Altogether, this boredom measure, alongside the two measurements of suicide and depression, were used to examine the hypothesis that was generated in the previous, bottom-up steps, regarding the direct and indirect links between boredom and suicide.

## Results

The analysis of the data was conducted via Pingouin – a commonly used statistics package written in Python.[38] The bottom-up analysis of the primary dataset from Facebook and the consecutive top-down analysis of both datasets are presented below.

**Bottom-up, topic-based analysis**

The stepwise regression model (step 3) of the entire set of topics (step 2) resulted in *four* topics that were significantly correlated with suicide risk [$R^2 = 0.037, F(4,1001) = 9.694, p < 0.0001$]. Table 1 presents the regression scores for each one of these four topics. The first topic that related to suicide [$\beta$=0.144, $t$(1005)= 3.3, $p$<0.005] consisted of boredom-related manifestations, as can be seen in the TF-IDF analysis and the response by ChatGPT (Step 4, Figure 2, Table S1, Table S2).

The second topic seemed to reflect the user's wish for something of value to him or her [$\beta$=0.142, $t$(1005)= 3.24, $p$<0.005]. Notably, most of the posts in this topic (93.5%) included an attached link, photo, or video, describing the specific needs of the participants (e.g., food, clothes, pet animals). The third topic was more difficult to interpret, but it seemed to address the participants' beliefs, or general view, about life and about their reality [$\beta$=0.133, $t$(1005)= 3.05, $p$<0.005]. Finally, the fourth topic [$\beta$=0.129, $t$(1005)= 2.97, $p$<0.005] addressed soft drugs and their legalization.

Table 1.

*Results from the stepwise regression that aimed to predict suicide risk from topics.*

| Predictor (topic) | $\beta$ coefficients | $t$ scores ($df$=1005) | $p$ value | 95% Confidence Intervals |
|---|---|---|---|---|
| Boredom | 0.1441 | 3.300 | 0.001 | [0.058, 0.230] |
| Wish for something | 0.1416 | 3.244 | 0.001 | [0.056, 0.227] |
| View about life | 0.1330 | 3.046 | 0.002 | [0.047, 0.219] |
| Drugs and legalization | 0.1299 | 2.977 | 0.003 | [0.044, 0.216] |
| Model Fit | $R^2$ = 0.037, $F$(4,1001) = 9.69, $p$ < 0.0001 | | | |

Note. See the third step in the Supplementary Material for the description of the input and output of the regression as well as the selection thresholds that were implemented in the analysis. This study focuses on the topic with the highest prediction value (i.e., boredom). Future studies are recommended to further examine the role of the remaining topics in the context of suicide research.

Figure 2.

*Word clouds based on the TF-IDF scores of the words in the four suicide-related topics.*

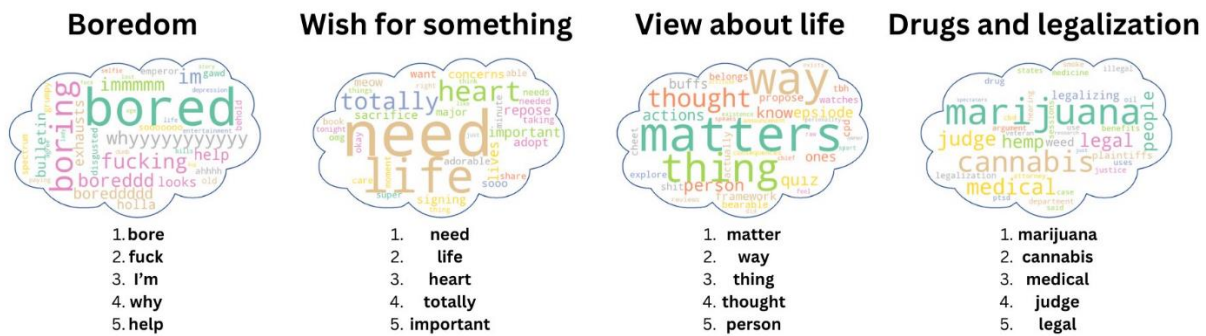

Note. The word clouds are ordered from left to right according to their prediction strengths (beta scores) of suicide risk. The size of the words in each cloud is proportional to their TF-IDF score. The 5 lemmatized words with the highest TF-IDF scores within each topic are presented beneath the clouds.

**Top-down hypothesis testing**

To validate the purely bottom-up finding regarding boredom, we calculated the correlations between the participants' scores on the boredom and the suicide scales in the secondary dataset (Step 5). This calculation indicated that the Disengagement factor was moderately correlated with the CSSRS ($r = .353$, $p < 0.001$; Figure 3a), thus providing a first confirmation of the hypothesis that boredom is linked to suicide behaviors.

To further characterize this link, we used the secondary dataset to conduct a path analysis,[38] which also considered depression, which has been linked before, both to suicide and to boredom (Introduction). As illustrated in Figure 3b, this analysis indicated that the direct path between boredom and suicide was not significant ($\beta = -.08$, SE=.048, $p = .093$), while the indirect path between boredom and suicide through the mediating variable of depression was significant and quite substantial ($\beta = .508$, SE = .048, p < .0001).

Finally, we replicated this path analysis using the original Facebook data and replaced the boredom questionnaire with the boredom topic (see also the Supplementary Material). This analysis resulted in two significant paths from boredom to suicide, an indirect path as documented in the previous analysis of the newly collected measures ($\beta = .046$, $SE = .024$, $p = .002$) as well as a direct path ($\beta = .021$, $SE = .040$, $p < .001$) that suggests that boredom itself can increase the risk for suicide (Figure 3). Moreover, in this dataset from Facebook, the bivariate correlation between the topic of boredom and suicide risk ($r = 0.185$, $p < 0.001$) was stronger than the correlation between boredom and depression ($r = 0.078$, $p < 0.05$). A similar pattern of results was evidenced when the proneness boredom scale was used (see the Supplementary Material).

Figure 3.

*Path analyses and bivariate correlations of Boredom, Depression, and Suicide in the Facebook data and the questionnaire data.*

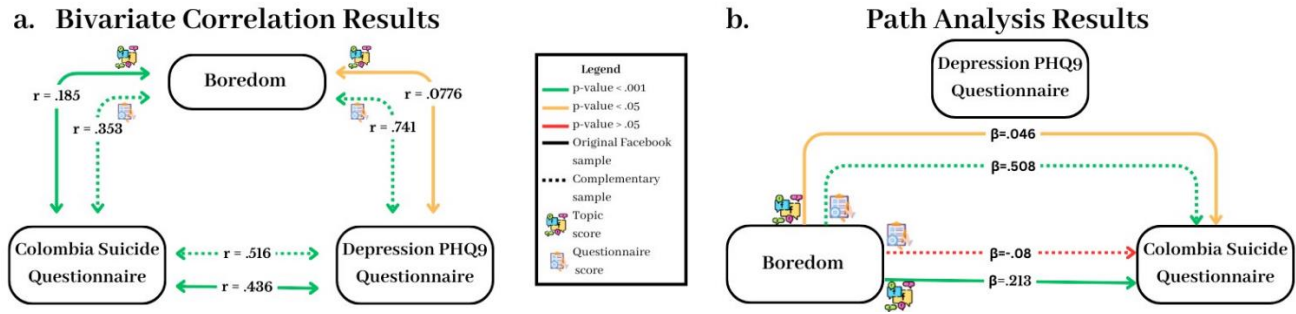

Note. The left graph (a) presents bivariate correlations between the three variables of these analyses, while the right graph (*b*) displays the results of the path analyses. The *inner dashed arrows* in both graphs represent the complementary data, and the *outer continues arrows* represent the original Facebook data. The color-coding on the graphs represents the statistical significance of the relationships: green for p-values < .001, yellow for p-values < .05, and red for p-values > .05. A supplemented cosine-similarity analysis that aims to further characterize the differences between the results from the primary Facebook dataset and the secondary questionnaire dataset is provided in the Supplementary Material.

## Discussion

This study, which consisted of a designated, 5-step research pipeline, aimed to harness the powerful LLMs for scientific discovery of risk factors for suicide. The first, purely bottom-up steps of the pipeline indicated that the topic that contributed the most to the prediction of suicide risk addressed experiences of boredom. The fifth, top-down step, which involved pre-defined hypotheses, resulted in two potential paths that link boredom to suicide behaviors:

The analysis of the secondary questionnaire-based dataset suggested that the least severe construct of boredom predicts the more severe moderator of depression, which in turn, predicts suicide ideation and behaviors. The analysis of the primary Facebook data indicated that boredom is linked directly to the risk of suicide, thus implying that boredom might serve as a maladaptive 'active ingredient' that is capable of triggering suicide ideation regardless of depression. Specifically, this mediation analysis of the Facebook data, as well as the observed superiority of the correlation between boredom and suicide over the correlation between boredom and depression (Figure 3), imply that the role of boredom in the creation and maintenance of suicide behaviors may be more crucial than what we might have thought based on the scarce and mostly outdated literature on this subject.

This last conclusion is noteworthy considering the existing research on boredom. As mentioned in the introduction, following the almost fully automated emergence of the topic of boredom, we used *APA PsycInfo* to search for the terms *boredom* and *depression* in any possible search field. This search yielded 46 results, of which 24 consisted of empirical studies, and only four were relevant to the hypothesized link between boredom and suicide:

A 30-year old research presented two surveys among Canadian and French adolescents, which yielded multiple links between suicide ideation and a range of risk factors, such as drug use, somatic complaints, and other self-perceived health problems, which included also a self-reported item regarding "bored/has a boring life".[19] A second, relatively old study, among 127 students reported a correlation ($r = 0.37$) between suicide risk and tedium, but the term 'tedium' may not serve as a good reflection of boredom as the authors defined it as an "experience of physical, emotional, and mental exhaustion".[21] A third, more recent PhD dissertation among prison inmates reported of multiple risk factors that were associated with suicide risk, including one of the items in the Psychopathy Checklist-Revised scale (PCL-R), which addressed the inmate's need for stimulation and his proneness to boredom.[20] Finally, a fourth study that followed 31 depressed patients for a week, reported that suicide ideation was preceded by feelings of tension, sadness, previous suicide ideation, and boredom.[22]

Together, the findings from this literature review, which indicated that boredom did not receive sufficient research attention in the context of suicide, and the findings from the current study emphasize the need to keep investigating the potential harmful role of boredom. It is noted that we do not argue that the current study provides a definite proof for a direct pathway between boredom and suicide. However, we strongly recommend that further research will aim to explore this direction, especially considering the ambiguity of the concept of boredom and its intertwined relationships with depression.

Despite the fact that all humans experience boredom from time to time,[39,40] the term 'boredom' and its distinctiveness from similar psychological constructs, such as anhedonia, apathy, and emptiness, are not consistently defined in the literature.[41] Correspondingly, a variety of theories were proposed over the years to explain this experience and to track its etiology, ranging from attention-based and arousal paradigms[42] to existential and psychodynamic theories.[43] Indeed, some evolutionary-oriented theories emphasized the potentially positive role of boredom, which could encourage people to seek change and engage in creative and meaningful activities.[44,45] However, in many cases, the experience of boredom seems to trigger negative emotions[46] and even psychopathologies,[47] including mainly depression.

The developers of the most common boredom proneness scale reported that it demonstrated moderate-size correlations ($0.47 \leq r \leq 0.52$) with three validated measurements of depression.[17] These strong correlations were expected, according to the authors, because boredom and depression share "overlapping symptomology" (p. 11). Similarly, the developers of the brief State Boredom Measure (SBM) reported that all of its eight items correlated positively with depression ($0.19 \leq r \leq 0.59$).[18] These psychometric data alongside the aforementioned theoretical statements about "overlapping symptomology" bring forth an unsettling discriminant validity issue whereby boredom is not clearly differentiated from depression.[45]

Indeed, some theorists aimed to pinpoint internal psychological aspects that differentiate boredom tendencies from pure depression.[48,49] These aspects might include *external ambivalence* in which bored persons typically blame others for their failures rather than directing anger towards themselves as is common in depression. They may also be characterized by *passive avoidance* and tend to avoid work or evade responsibility rather than keep trying while failing to win the love and admiration of others as often is the case with depression.[48] However, in practice, the phenomenological overlap between these two concepts of boredom and depression makes the research on the implications of boredom rather difficult. This discriminant validity problem may explain why the results from the path analysis of the questionnaire data were not identical to the results from the Facebook data. In the questionnaire data, it is possible that the concept of boredom was not differentiated sufficiently from the concept of depression, thus preventing us from isolating its direct impact on suicide.

**Limitations**

This study has several limitations. Firstly, the social media data of this study was extracted from relatively active users of Facebook,[9] thus limiting the generalizability of its bottom-up findings. Users who suffer from depression (a close predictor of suicide), for example, might be less active, or hesitant to share their thoughts or emotions online,[50] and therefore be less represented in this sample of Facebook users. Secondly, despite the similarity between the Facebook-based topic of boredom and the questionnaire-based concept of boredom (Figure S1), the two psychological constructs may not be identical. This characteristic may explain the different patterns of results, but it also limits our ability to provide a perfect confirmation of the bottom-up findings through the top-down analyses. Lastly, the observational nature of the two datasets used in this study limits our ability to propose causal oriented conclusions, thus emphasizing the importance of conducting further studies on this topic, preferably using longitudinal research designs.

**Clinical implications**

Considering these limitations, the findings of the current study should be interpreted with caution. Nevertheless, the purely bottom-up nature of the key finding of this study emphasizes the potential hazard in continuous, burdening experiences of boredom. Although boredom is a very common experience, its negative implications may be undervalued, both in the literature and in the clinical field. For some people, boredom may indeed feel like a relatively benign experience, or even a positive experience as it could lead to creative and novel ideas and activities, as well as meaningful positive change.[44,45] However, for many people, boredom is perceived as a fairly aversive experience,[40] and our findings suggest that it might even lead a person to engage in dangerous suicidal behaviors.

The mechanism that might explain this hazard has not been the focus of the current study. However, based on the existing literature on boredom, one potential explanation may relate to the increased risk for self-harm behaviors in boredom experiences, which in turn, might evolve into actual suicidal behavior. In the well-cited series of studies on "the challenges of the disengaged mind", a seemingly harmless and easy task of spending 6-15 minutes doing nothing was experienced by most participants as an inherently unpleasant mission, to the point that many participants (especially male participants) chose to give themselves an electric shock.[51] In a way, for these individuals, an adverse stimulation was preferred to no stimulation at all.

Indeed, self-harm behaviors are typically understood as a (non-adaptive) method to regulate unwanted negative emotions.[52,53] However, there seems to be convincing evidence that self-harm behaviors can also result simply from the person's wish to avoid tedious experiences and disrupt their burdening monotony. Several experiments on this topic, which included benign control groups as well as groups that were induced with negative emotions, found that the 'active psychological ingredient' that led participants to self-administer electric shocks was boredom.[54-56] This 'dark side' of boredom has even led researchers to reconsider its role in the self-harming behaviors that characterize the common Borderline Personality Disorder.[57]

A second explanation for the observed link between boredom and suicide may be the enmeshment between boredom and deeper existential experiences, such as emptiness[58] and lack of meaning in life,[59] which are recognized as substantial risk factors of suicide.[60] To our knowledge, these subjective, amorphic, and even mysterious human experiences typically do not receive sufficient space in the context of evidence-based treatments for suicide, such as interpersonal psychotherapy (IPT)[61] and Cognitive-Behavioral Therapy (CBT).[62] We therefore conclude that boredom should be further considered, both in theory as argued above, and in practice, during suicide prevention programs and therapies.

This conclusion is noteworthy considering the current state of the literature on suicide. Suicide research, as described in the introduction, typically revolves around the same risk factors,[5] and this study demonstrated that the integration of AI methods could lead to the exposure of under-recognized risks, such as boredom. Boredom, as documented in our thorough review of the literature above, has rarely been investigated directly within the context of suicide, and the current AI-inspired research allowed us to identify it and further explore it using complementing top-down analysis. We therefore join previous calls to combine AI tools in suicide research,[8,63] in a careful and responsible manner, as these tools can improve our understanding of suicide, and perhaps direct us to craft more clinically tailored treatments to individuals at risk.

Bored to Death: Artificial Intelligence Research Reveals the Role of Boredom in Suicide Behavior

<div align="center">**Supplementary Material**</div>

**Primary dataset from Facebook**

The complete description of the Facebook dataset used in the current study is available in our previous publication that focused on prediction of suicide risk, rather than on understanding of the suicide phenomenon.[1] However, two characteristics differentiate between the current dataset and the previous one. First, the current sample ($N = 1,006$) included four additional participants whose Facebook activity could not be investigated with the (less advanced) language model that was used in our previous study. Second, the number of Facebook postings in the current dataset ($N = 228,052$) was larger than the number of postings in the previous dataset as it included also (textual) postings that were attached to videos, images, and internet links. This is in contrast to our previous study, which focused on standalone textual postings only.

<div align="center">**5-step research pipeline**</div>

As mentioned in the main article, the procedures of the study consisted of a 5-step research pipeline (Figure 1, main article). Below is a detailed description of the five steps.

**Step 1. Representing the Facebook texts**

In the first step, we extracted numeric representations of the Facebook texts we collected, using a popular Deep Language Model (DLM) named Sentence-BERT (SBERT).[2] SBERT is a variant of the well-established BERT model,[3] and it is specifically suitable for generating representations of short texts, such as social media postings.

In addition to the two objectives that served the development and training of the original BERT model ("masked language model" and "next sentence prediction"),[3] SBERT has been further optimized to generate sentence representations that reflect similarities between sentences. These representations appear in the form of numerical, multidimensional vectors known as "embedding vectors", which have been demonstrated to capture many linguistic and semantic aspects of the language.[2] In this study, we utilize the "base" version of SBERT (via the implementation of https://huggingface.co/sentence-transformers), which takes as input the entire sentence and provides a 768-dimensional vector.

These vectors are also known as 'embedding vectors' and previous studies showed that they are able to capture many linguistic and semantic aspects of natural language.[2] Altogether, the representation process resulted in 228,052 vectors, one for each Facebook text.

## Step 2. Clustering the text representations into topics

In the second step, we applied a clustering algorithm with the goal of organizing the different posts, which were now represented numerically, into groups that ideally capture meaningful 'topics' (we refer to the resulted clusters as topics because the spatial proximity of the representations is assumed to reflect proximity in the semantic meaning of the posts). This is achieved by projecting posts into a numerical space with semantic significance. Consequently, posts sharing similar subjects or themes tend to belong to the same cluster, which can be labeled with the common topic represented by the posts within it. Note that there are many other traditional methods for "topic modeling" (e.g., LDA: Latent Dirichlet Allocation)[4] that are based on statistical distributions of words/n-grams in the text. However, in this study we leveraged the power of the more rich and fine-grained embedding representations that capture the entire context of the words and not just the words themselves. Furthermore, the clustering method ensures that each post is assigned to only one topic, a property that is crucial for subsequent analysis (see Step 3).

The clustering algorithm that was used for this task was HDBSCAN.[5] This algorithm applies a hierarchical density-based clustering method and automatically determines the optimal number of clusters (minimum cluster size = 25). Before applying the algorithm, the embedding representation vectors were dimensionally reduced from 768 to 10 dimensions using the Uniform Manifold Approximation and Projection (UMAP) algorithm.[5] Altogether, this process yielded 771 topics, of which we removed 60 topics that were shared by less than ten different users, in order to avoid esoteric topics that are used only by very few. The number of posts in each topic ranged from 25 to 4956 ($M =$ 96.9, $SD =$ 259.9 ).

## Step 3. Identifying the topics that are most related to suicide risk

In the third step, we aimed to filter out irrelevant topics and identify the topics that are mostly correlated with suicide risk. To achieve this aim, we implemented a stepwise regression model[6] that predicts the users' suicide risk score (i.e., the CSSRS scores, which ranged from 0 to 6) based on their proportional use of each of topic. More specifically, for each user $u$, we created a numerical topic distribution vector $d_u$:

$$d_u = \left( p_{topic\ 1}, p_{topic\ 2}, \dots, p_{topic\ n} \right)$$

Where $n$ is the number of topics and $p_{topic\ i}$ is the proportion of posts user $u$ published in topic $i$. These vectors were fed as input to the regression model, with each topic considered as a potential independent variable. To determine a manageable number of topics and facilitate interpretation, we introduced strict thresholds in our stepwise algorithm. The forward selection (inclusion) threshold was set to p-values < 0.003 and the backward selection (exclusion) threshold was set to p-values > 0.005. This step yielded a small set of topics that can be viewed as the most significant predictors of suicide risk.

## Step 4. Analyzing the thematic content of the suicide-related topics.

In the fourth step, we interpreted the resulting set of topics from the previous step, using three methods: a manual inspection of posts sampled from the topics by three of the authors, a 'consultation' with the currently popular LLM of ChatGPT[7] (which was asked to propose a shared topic for each list of posts, see Table S1), and a well-established analysis named TF-IDF (Term Frequency – Inverse Document Frequency).[8] TF-IDF provides statistical measurements to weight words according to their relevance within each document (or 'topic,' in this case), by taking into consideration both the frequencies of words in the topic and their overall prevalence in the other topics. This method allowed us to identify the most important words of each topic and thereby to reveal its main theme (Table S2).

The convergence of evidence obtained through these three methods for analyzing thematic content led us to assign a general conceptual label to each topic. Note that we could not provide real examples of postings that were classified into the various topics due to privacy considerations.

Table S1.

*A chatGPT description of the posts in each of the significant buttom-up topics[7]*

| Topic | ChatGPT topic decription |
|---|---|
| Boredom | The common topic of all the posts you provided is boredom. |
| Wish for something | The topic of the posts appears to be expressing a desire or need for various things or experiences in the authors' lives. These could range from material objects to emotional needs, such as love or support. The specific items or experiences mentioned in the posts vary widely, from tangible items like books and products to abstract concepts like personal growth and emotional connections. |
| View about life | The topic of the following posts is varied, covering a range of subjects and personal thoughts. There isn't a single overarching topic, but rather a collection of individual thoughts, observations, and statements shared by different individuals on social media. These posts touch on subjects like relationships, self-perception, uniqueness, opinions on various matters, personal experiences, and more. |
| Drugs and legalization | The topic of the following posts is primarily about marijuana, including discussions about its legalization, medical benefits, its legal status, and related political and social issues. |

Note. The prompt that was used to receive a response from GPT was: "what is the mutual topic of the following posts: [list of posts that were clustered together in the second step of the research pipeline]".

Table S2.

*Top 10 TF-IDF words in each of the significant buttom-up topics*

|  | Boredom | Wish for something | View about life | Drugs and legalization |
|---|---|---|---|---|
| 1 | Bore (2.85) | Need (0.94) | Matter (0.22) | Marijuana (0.23) |
| 2 | Fuck (0.16) | Life (0.80) | Way (0.20) | Cannabis (0.13) |
| 3 | Im (0.16) | Heart (0.15) | Thing (0.18) | Legal (0.08) |
| 4 | Why (0.13) | Totally (0.12) | Thought (0.17) | Medical (0.06) |
| 5 | Help (0.12) | Important (0.12) | Person (0.13) | Judge (0.05) |
| 6 | Exhausts (0.11) | Repose (0.11) | Quiz (0.12) | Hemp (0.04) |
| 7 | Bulletin (0.10) | Concern (0.09) | Know (0.11) | People (0.03) |
| 8 | Look (0.09) | Adopt (0.09) | Action (0.10) | Weed (0.03) |
| 9 | Emperor (0.09) | Sacrifice (0.08) | Epsiode (0.09) | Plaintiff (0.03) |
| 10 | So (0.09) | Sign (0.08) | One (0.08) | Drug (0.03) |

Note. The words in each column are ordered by their TF-IDF scores (which are presented in parenthesis).

**Step 5. Using a top-down hypothesis testing to further examine the strongest suicide-related topic (i.e., boredom).**

To validate and further examine the results from the previous steps, we re-examined them using more conventional, top-down hypothesis testing methods. Note that the a-priory hypotheses needed to implement this top-down step could have only been formulated after we obtained the results of the previous steps, which suggested that boredom plays a role in the creation or maintenance of the suicide risk (Results).

In this step, as mentioned in the main article, we collected a new sample of 1,062 participants from MTurk and asked them to complete three psychological measurements: (1) the Columbia Suicide Severity Rating Scale (CSSRS),[9,10] (2) the 9-item Patient Health Questionnaire (PHQ-9) measuring depression,[11-13] and (3) the Multidimensional State Boredom Scale (MSBS).[14] This last scale consists of 28 items that aims to assess five theoretical dimensions of boredom. Notably, the central dimension of this scale, the Disengagement dimension that has the largest number of items and the highest factor loading, reflected the topic of boredom that emerged from the primary Facebook data (see below).

### Similarity analysis

As a complementary analysis we wished to evaluate the extent to which the bottom-up topic of boredom (which was extracted using the clustering algorithm that was applied to the LLM-based representations of the Facebook posts) is indeed related to the more formal concept of boredom, as might be extracted from the validated questionnaire that measures it. To this end, we used SBERT (see the Methods in the main article) to produce vector representations for the items that comprise the disengagement factor of the Multidimensional State Boredom Scale.[14] In addition, we produced representations of the items of the two other common psychological scales that could serve as control scales for the analysis below – the previously mentioned PHQ-9, which measures depression,[13] and the short version of the well-known Big Five Inventory.[15] In addition to the straightforward comparison with the boredom questionnaire, we were interested in the connection to depression, as the psychological literature draws meaningful connections between boredom and depression (see the introduction section of the main article). Furthermore, we sought to gauge the validity of our findings by employing a broad-spectrum personality assessment, the Big Five questionnaire, as a general measure.

Following this procedure, we computed the similarity score of these scales with the SBERT representation vectors of the postings that were classified to the boredom topic. To calculate this similarity score, we computed the similarity between each sentence in a post, since sentence level is more compatible to match a questionnaire item, and each item in the questionnaire. The final score for each post was obtained by selecting the highest similarity score between a sentence and a questionnaire item. To compute the overall similarity score between the questionnaire and a topic, we took the average of all the similarity scores calculated between the questionnaire and the posts of that topic. When comparing the similarities between the scores of the different scales, we revealed that the topic of boredom was indeed closer in space to the boredom scale than to the two control scales (Figure S1).

We completed this analysis with a second analysis of 217 random postings that were sampled from the entire Facebook data. We used SBERT to assign representations to these random postings (which matched in their number to the number of postings in the boredom topic) and compared them to the representations of the aforementioned three scales. The random sample was most similar to the Big Five Inventory. We also performed a statistical hypothesis testing, where the null hypothesis is that the similarity between the boredom topic and the random sample to each questionnaire are equal. This comparison revealed that the similarities were significantly different in the cases of the boredom questionnaire ($p < .001$) and the depression questionnaire ($p < .001$) but not in the Big Five Inventory ($p > .05$). It could be concluded then that the topic of boredom reflects, even if not completely, the more formal construct of boredom, as captured by the standard MSBS questionnaire. In other words, this analysis illustrates how LLMs are indeed relevant for the task of identifying hidden psychological constructs.

Our rigorous examination sought to determine the extent to which topic-based boredom posts align with boredom and depression questionnaire items. The results of this analysis present a substantial similarity between boredom-topic posts and items within boredom and depression questionnaires. Nonetheless, further investigation is warranted to comprehensively elucidate the connections and distinctions between topic-based boredom and boredom measured by traditional questionnaires.

Figure S1

*Comparing the representations of the topic of boredom with representations of the items in the boredom scale and in two control scales.*

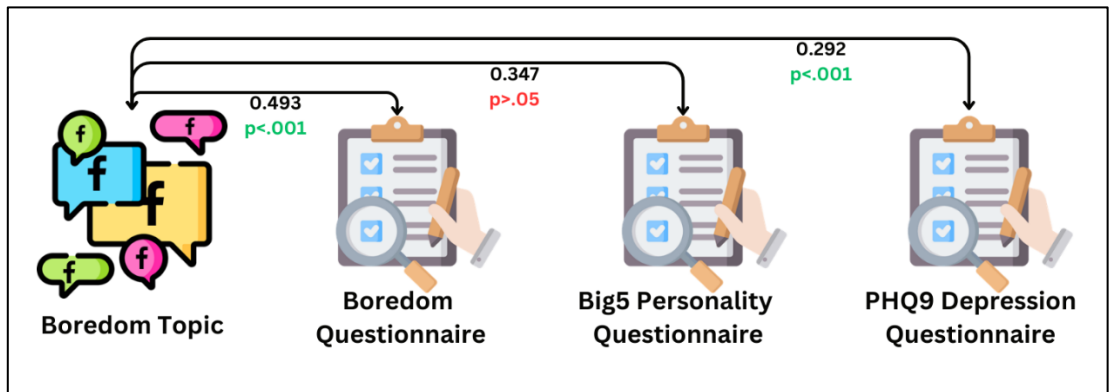

Note. The p-values of this Figure were obtained from a t-test hypothesis testing, comparing the average similarity scores between the topic and the questionnaires to those of randomly selected posts and the questionnaires.

## Path analysis of the Facebook data

The path analysis presented in the main article addressed three psychological constructs: suicide risk, depression, and boredom. The measures used to obtain the first two constructs in the primary Facebook dataset were the same as the measures implemented in the secondary questionnaire dataset (i.e., the CSSRS[16] and the PHQ-9[13]). The third measure of boredom, in contrast, was obtained differently in the two datasets. In the primary Facebook dataset, the boredom scores were assigned to participants based on their usage of the boredom topic, while in the secondary questionnaire dataset the boredom scores were assigned to the participants based on their responses to the boredom scale (i.e., the MSBS).

This difference created a significant gap between the two datasets. While the secondary questionnaire dataset included boredom scores for all the participants (as all were required to complete the boredom questionnaire), in the primary Facebook data, only 3% of the sample (35 of 1,006 participants) had postings that were assigned to the topic of boredom by the clustering algorithm that was used in the study. This low percentage occurred because the clustering algorithm allowed the assignment of only one topic to each vector representation (Facebook post). In light of this gap, before we implemented the path analysis of the Facebook data, we first enriched and extended the topic of boredom, so more Facebook users could be given boredom scores.

To this end, we created two pools of potentially relevant postings. The first pool consisted of the closest 500 posts to the centroid of the original boredom topic based on the spatial proximity of the vector representations. The reasoning behind this process it that these 500 posts would have been assigned to this cluster with a more tolerant algorithm. The second pool consisted of postings that had explicit boredom-related language (i.e., words with the lingual root of boredom that might have not been classified originally to the boredom topic since the leading topic in these postings required their classification to another topic). We then reviewed the postings of these two pools manually and integrated postings that had clear boredom related content within the original topic of boredom. Illustrative postings of boredom experiences that emerged in this enrichment process are: "I'm mentally drained!", "Why Saturday is so boring?", and "I am bored again, there is no one to talk to". Altogether, we enriched the boredom topic with 53 postings from the first pool and 123 postings from the second pool.

Following this enrichment process, the total number of participants receiving a boredom score larger than zero was 121 (12%), i.e., 86 more participants than the original boredom coding. These enriched boredom scores were then used to compute the path analysis between boredom, depression, and suicide (see the results section of the main article).

## Complete statistics for all path analyses

### Analysis of the primary Facebook data using the Boredom Topic

| Path | coef | se | t | p | Confidence Interval | R^2 |
|---|---|---|---|---|---|---|
| PHQ-9 ~ Boredom Topic | Boredom Topic 0.078 | 0.031 | 2.46 | 0.01 | [0.016, 0.139] | 0.006 |
| CSSRS ~ PHQ-9, Boredom Topic | Boredom Topic 0.213 | 0.040 | 5.39 | 0.000 | [0.136, 0.291] | |
| | PHQ-9 0.597 | 0.040 | 15.08 | 0.000 | [0.519, 0.674] | 0.213 |

| | Effect | SE | p | Confidence Interval |
|---|---|---|---|---|
| Total | 0.260 | 0.043 | 0.000 | [0.174, 0.345] |
| Direct | 0.213 | 0.040 | 0.000 | 0.136, 0.291] |
| Indirect | 0.046 | 0.024 | 0.002 | [0.006, 0.101] |

### Analysis of the secondary, questionnaire data using the Multidimensional State Boredom Scale (MSBS)[14]

| Path | coef | se | t | p | Confidence Interval | R^2 |
|---|---|---|---|---|---|---|
| PHQ-9 ~ MSBS | MSBS 0.741 | 0.020 | 35.95 | 0.000 | [0.701, 0.782] | 0.549 |
| CSSRS ~ PHQ-9, MSBS | MSBS -0.080 | 0.048 | -1.68 | 0.093 | [-0.173, 0.013] | |
| | PHQ-9 0.687 | 0.048 | 14.43 | 0.000 | [0.593, 0.780] | 0.269 |

| | Effect | SE | p | Confidence Interval |
|---|---|---|---|---|
| Total | 0.429 | 0.035 | 0.000 | [0.360, 0.497] |
| Direct | -0.080 | 0.048 | 0.093 | [-0.173, 0.013] |
| Indirect | 0.508 | 0.051 | 0.000 | [0.412, 0.612] |

**Analysis of the secondary, questionnaire-based dataset using the Boredom Proneness Scale (BPS)[17]**

| Path | Coefficient | se | t | p | Confidence Interval | R^2 |
|---|---|---|---|---|---|---|
| PHQ-9 ~ BPS | BPS 0.672 | 0.023 | 29.57 | 0.000 | [0.628, 0.717] | 0.452 |
| CSSRS ~ PHQ9, BPS | BPS -0.042 | 0.043 | -0.97 | 0.331 | [-0.127, 0.043] | |
| | PHQ-9 0.656 | 0.043 | 15.19 | 0.000 | [0.571, 0.740] | 0.267 |

| | Effect | SE | p | Confidence Interval |
|---|---|---|---|---|
| Total | 0.399 | 0.032 | 0.000 | [0.329, 0.468] |
| Direct | -0.042 | 0.043 | 0.331 | [-0.127, 0.043] |
| Indirect | 0.441 | 0.043 | 0.000 | [0.356, 0.527] |